\documentclass[a4paper,11pt]{article}

\usepackage{amssymb,amsmath,array}
\usepackage{hyperref}
\usepackage{bm}

\usepackage[usenames,dvipsnames,svgnames]{xcolor}
\usepackage{tango-colors}

\usepackage{tikz}
\usetikzlibrary{arrows}
\usetikzlibrary{shapes}
\usetikzlibrary{shapes.misc}
\usetikzlibrary{calc}
\usetikzlibrary{backgrounds}

\usepackage{pgfplots}
\pgfplotsset{compat=1.13}
\usepgfplotslibrary{fillbetween}
\usepgfplotslibrary{groupplots}

\usepackage[a4paper, left=3cm, right=2.5cm, top=3cm, bottom=3cm]{geometry}

\usepackage{authblk}
\usepackage{natbib}
\usepackage{doi}

\usepackage[utf8]{inputenc}
\usepackage[T1]{fontenc}
\usepackage[english]{babel}
\usepackage{csquotes}

\renewcommand{\cite}{\citep}

\newcommand{\R}{\mathbb{R}}

\newcommand{\deriv}[1]{\frac{\partial}{\partial #1}}

\newcommand{\jac}{J}
\newcommand{\eig}{\lambda}

\newcommand{\quali}{q}
\newcommand{\thresh}{\theta}
\newcommand{\fun}{f}
\newcommand{\scr}{g}
\newcommand{\lbl}{y}

\newcommand{\prob}{P}
\newcommand{\dens}{p}
\newcommand{\mean}{\mu}

\newcommand{\nDens}{\mathcal{N}}
\newcommand{\nCum}{\Phi}
\newcommand{\nMean}{\mu}
\newcommand{\nDev}{\sigma}
\newcommand{\expec}{\mathbb{E}}

\newcommand{\parrate}{k}

\newcommand{\npop}{m}
\newcommand{\persidx}{i}

\newcommand{\naccpt}{n}
\newcommand{\group}{C}
\newcommand{\ingrp}{c}
\newcommand{\outgrp}{{\neg c}}

\newcommand{\gap}{\eta}

\newcommand{\timeidx}{t}

\newcommand{\leak}{\alpha}
\newcommand{\win}{\beta}

\newbool{arxiv}
\setbool{arxiv}{true}

\author{Benjamin Paaßen}
\author{Astrid Bunge}
\author{Carolin Hainke}
\author{Leon Sindelar}
\author{Matthias Vogelsang}
\affil{CITEC Center of Excellence, Bielefeld University\thanks{Funding by the CITEC center of excellence (EXC 277) is gratefully acknowledged.}}

\title{Dynamic fairness -- Breaking vicious cycles in automatic decision making}

\date{Preprint of the ESANN 2019 paper \citet{Paassen2019ESANN} as provided by the authors.
The original can be found at the \href{https://www.elen.ucl.ac.be/esann/proceedings/electronicproceedings.htm}{ESANN electronics proceedings page}.}

\begin{document}

\maketitle

\pagestyle{myheadings}
\markright{Preprint of \citet{Paassen2019ESANN} provided by the authors.}

\begin{abstract}
In recent years, machine learning techniques have been increasingly applied in
sensitive decision making processes, raising fairness concerns.
Past research has shown that machine learning may reproduce
and even exacerbate human bias due to biased training data or flawed model
assumptions, and thus may lead to discriminatory actions.
To counteract such biased models, researchers have proposed multiple mathematical
definitions of fairness according to which classifiers can be optimized.
However, it has also been shown that the outcomes generated by some fairness
notions may be unsatisfactory.

In this contribution, we add to this research by considering decision making processes
in time. We establish a theoretic model in which even perfectly accurate classifiers which adhere
to almost all common fairness definitions lead to stable long-term inequalities due to
vicious cycles.
Only demographic parity, which enforces equal rates of positive decisions across groups,
avoids these effects and establishes a virtuous cycle, which leads to perfectly
accurate and fair classification in the long term.
\end{abstract}

Automatic decision-making via machine learning classifiers carries the promise of
quicker, more accurate, and more objective decisions
because automatic mechanisms do not foster animosity against any group
\cite{WhiteHouse2016,ONeil2016}. Yet, machine learning systems can indeed reproduce
and exacerbate bias that is encoded in the training data
or in flawed model assumptions \cite{WhiteHouse2016,ONeil2016,Corbett2018,Dwork2012}.
For example, the COMPAS tool, which estimates the risk of recidivism of defendants
in the US law system prior to trial, has been found to have higher rates of false
positives for Black people compared to white people and has thus been called unfair
\cite{Angwin2016}. Similarly, a tool developed by Amazon to rate the résumés of job 
applicants assigned higher scores to men compared to women because successful
applicants in the past had mostly been male \cite{Dastin2018}. Finally,
multiple machine-learning-based credit scoring systems have emerged that reproduce
historical biases and systematically assign lower credit scores to members of
disenfranchised minorities \cite{ODywer2018}.

In general, we consider scenarios where individuals
$\persidx \in \{1, \ldots, \npop\}$ in some population of size $\npop$ apply for some positive
outcome, such as a pre-trial bail, a job, or a loan, and a gatekeeper institution decides whether
to grant that outcome, with the interest of accepting only those individuals who will
\enquote{succeed} with that outcome, e.g.\ not commit a crime, succeed in their job for the company,
or pay back a loan. To make that decision, the institution employs a binary classifier
$\fun : \{1, \ldots, \npop\} \to \{0, 1\}$ that predicts whether to grant the outcome, i.e.
$\fun(\persidx) = 1$, or not, i.e.\ $\fun(\persidx) = 0$. Now, let $\lbl_\persidx \in \{0, 1\}$ denote
whether an individual will succeed ($\lbl_\persidx = 1$) or not ($\lbl_\persidx = 0$). Then, the
aim of the classifier is to maximize the share of the population where $\fun(\persidx) =
\lbl_\persidx$.

In our examples, we care how a certain \emph{protected group} $\group$ is treated compared to everyone
else. In general, we assume these protected groups to be pre-defined by society, e.g.\
via the EU charter of fundamental rights, which forbids discrimination based on sex, race, color,
ethnic or social origin, religion, political opinion, and several other features \cite{EU2012}.
Formally, let $\group \subseteq \{1, \ldots, \npop\}$ be a protected group, let
$\npop_\ingrp := |\group|$, let $\neg \group := \{1, \ldots, \npop\} \setminus \group$, and let
$\npop_\outgrp := |\neg \group|$. Then, the fairness notion corresponding to our last two
examples is \emph{demographic parity}, which requires that the rate of positive decisions
is equal across groups, i.e.\ $\sum_{\persidx \in \group} \frac{\fun(\persidx)}{\npop_\ingrp} =
\sum_{\persidx \in \neg \group} \frac{\fun(\persidx)}{\npop_\outgrp}$ \cite{Dwork2012,Zliobaite2017}.

Multiple authors have criticized demographic parity because it decreases accuracy if the base rate of
successful people is different across groups \cite{Corbett2018,Zliobaite2017,Hardt2016}. Accordingly,
Hardt et al.\ have proposed the notion of \emph{equalized odds} which only requires an equal rate of
positive decisions among the people who will succeed and the people who will not succeed \cite{Hardt2016},
which corresponds to the fairness notion in the COMPAS example \cite{Angwin2016}.

In addition distributive justice considerations, several authors have proposed notions of
due process, in the sense that any classifier should be considered fair which performs decisions
in a fair way \cite{Grgic2016}. In particular, several authors have argued that classifiers should
not use features that code the protected group directly or indirectly
\cite{ONeil2016,Dwork2012,Grgic2016,Kilbertus2017,Kusner2017}. Alternatively, Corbett and Goel have
proposed a two-step classification process. First, a function
$\scr : \{1, \ldots, \npop\} \to \R$ assigns a risk score to each individual, which should increase
monotonously with the probability to be successful, i.e. $\scr(\persidx) = \sigma(\prob(\lbl_\persidx = 1))$
for some monotonous function $\sigma$ (a property also called \emph{calibration} \cite{Liu2018}).
Second, the actual classifier only threshold the risk score, i.e.\ $\fun(\persidx) = 1$
if $\scr(\persidx) \geq \thresh$ and $\fun(\persidx) = 0$ otherwise for some fixed threshold
$\thresh \in \R$, thus holding everyone to the same standard \cite{Corbett2018}.

In this contribution, we argue that even if a classifier is perfectly accurate and is fair
according to all fairness notions except demographic parity, we may still obtain undesirable
long-term outcomes. To do so, we establish a simple dynamical system which assumes that positive
classifier decisions have positive impact on the future success rate of a group, which in turn
leads to a higher chance for positive classifications and so on. We show that this positive
feedback loop implies stable equilibria where a protected group receives no positive
decisions anymore. We also show that imposing demographic parity breaks this feedback loop
and introduces a single, stable equilibrium which exhibits perfect accuracy, equality, and
fairness according to all notions.

Our model is inspired by prior work of O'Neil, who has investigated existing automatic decision
making systems and found positive feedback loops which disadvantage protected groups
\cite{ONeil2016}. However, O'Neil did not provide a theoretic model. Further,
our work is related to prior research by Liu et al., who have analyzed one-step dynamics in a
credit scoring scenario \cite{Liu2018} but did not consider long-term outcomes. Third, Hu and
Chen have previously analyzed a detailed economic model of the labor market, including long-term
dynamics \cite{Hu2018} and found that demographic parity leads to a desirable equilibrium.
Finally, Mouzannar et al.\ generalized this work simultaneously and independently to us and analyzed a wide range of scenarios where
acceptance decisions influence future qualifications \cite{Mouzannar2019}. Our work is similar
to theirs, but we use a different model assuming continuous qualification variables,
fixed institutional resources, and specific dynamics, which enables us to derive stronger conclusions.

\section{Model}

In our model, we assume that every individual $\persidx$ has an objective risk score
$\quali_\persidx^\timeidx$ at time $\timeidx$ which is drawn from an exponential
distribution%
\footnote{Note that our qualitative results can be generalized to other distributions, such as Gaussian or Pareto.
We select the exponential distribution here because it only has a single parameter and thus is easier to analyze.
\ifboolexpr{bool {arxiv}}{You can find the full analysis in Appendix~\ref{appendix}.}%
{You can find the full analysis in the appendix at \url{https://arxiv.org/abs/1902.00375}.}
}
with mean $\mean_\ingrp^\timeidx$ if $\persidx \in \group$ and with mean
$\mean_\outgrp^\timeidx$ otherwise. Further, we assume that the $\naccpt \leq \npop$ people with
the highest score in each iteration are the ones which will be successful, i.e.\ $\lbl_\persidx
= 1$ if and only if $\quali_\persidx^\timeidx$ is among the top $\naccpt$ at time $\timeidx$.
Accordingly, we obtain a perfectly accurate classifier if we use the scoring function $\scr^\timeidx(\persidx)
= \quali_\persidx^\timeidx$ and set the decision threshold $\thresh^\timeidx$ such that exactly the top
$\naccpt$ scores are above or equal to it. Note that our hypothetical classifier conforms to equalized odds
because there are no misclassifications \cite{Hardt2016}, fulfills the calibration, threshold, and accuracy
requirements of Corbett and Goel \cite{Corbett2018}, and does not need access to the group label,
neither directly nor indirectly, thus conforming to all due process notions of fairness
\cite{ONeil2016,Dwork2012,Grgic2016,Kilbertus2017,Kusner2017}.

We estimate the overall number of people who receive a positive classification
inside and outside the protected group via the expected values
$\expec[\sum_{\persidx \in \group} \fun(\persidx)] = \npop_\ingrp \cdot \int_{\thresh^\timeidx}^\infty
\frac{1}{\mean_\ingrp^\timeidx} \cdot \exp(-\frac{\quali}{\mean_\ingrp^\timeidx}) d\quali = \npop_\ingrp \cdot \exp(-\frac{\thresh^\timeidx}{\mean_\ingrp^\timeidx})$
and $\expec[\sum_{\persidx \in \neg \group} \fun(\persidx)] = \npop_\outgrp \cdot \exp(-\frac{\thresh^\timeidx}{\mean_\outgrp^\timeidx})$%
\footnote{We consider each classifier decision as a Bernoulli trial with success probability $\prob = \int_{\thresh^\timeidx}^\infty
\frac{1}{\mean_\ingrp^\timeidx} \cdot \exp(-\frac{\quali}{\mean_\ingrp^\timeidx}) d\quali$, yielding
a binomially distributed random variable $\sum_{\persidx \in \group} \fun(\persidx)$ with
expected value $\npop_\ingrp \cdot \prob$ and variance $\npop_\ingrp \cdot \prob \cdot (1 - \prob)$.
Note that the variance gets close to zero if $\prob$ is small itself, such that the expected value
is a precise estimate for sufficiently small $\naccpt$.}.

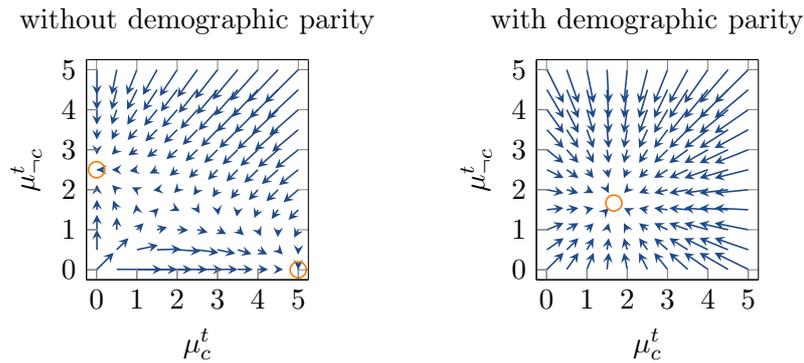
\begin{figure}
\begin{center}
\begin{tikzpicture}
\begin{groupplot}[
group style={%
group size=2 by 1, horizontal sep=3cm},
view={0}{90},
xmin=-0.25,xmax=5.25,
ymin=-0.25,ymax=5.25,
width=4.5cm,height=4.5cm,
xlabel={$\mean_\ingrp^\timeidx$},
ylabel={$\mean_\outgrp^\timeidx$},
xtick={0, 1, 2, 3, 4, 5},
ytick={0, 1, 2, 3, 4, 5}
]
\nextgroupplot[title={without demographic parity}]
\addplot3[skyblue3,
semithick,
quiver={u=\thisrow{u_alpha0.5_beta5},v=\thisrow{v_alpha0.5_beta5},scale arrows=0.5,},
-stealth] table[x=mu_C, y=mu_nC] {exponential_deltas.csv};
\draw[orange2, semithick] (axis cs:0,2.5) circle[radius=0.2];
\draw[orange2, semithick] (axis cs:5,0) circle[radius=0.2];
\nextgroupplot[title={with demographic parity}]
\addplot3[skyblue3,
semithick,
quiver={u=\thisrow{u_alpha0.5_beta5},v=\thisrow{v_alpha0.5_beta5},scale arrows=0.5,},
-stealth] table[x=mu_C, y=mu_nC] {exponential_dp_deltas.csv};
\draw[orange2, semithick] (axis cs:1.667,1.667) circle[radius=0.2];
\end{groupplot}
\end{tikzpicture}
\end{center}
\caption{An illustration of the dynamical system model from Equation~\ref{eq:exp_model} for a
population with $\npop_\ingrp = 100$, $\npop_\outgrp = 200$, $\naccpt = 50$ successful people,
leak rate $\leak = 0.5$, and score $\win = 5$. Equilibria are highlighted with circles.
Left: The model without demographic parity requirement, exhibiting undesirable stable equilibria at the
coordinate axes. Right: The model with demographic parity, exhibiting a single stable equilibrium on
the diagonal.}
\label{fig:systems}
\end{figure}

We finally assume that the mean for a group improves with a higher rate of positive classifier
decisions in the previous time step according to the following equation.
\begin{equation}
\begin{pmatrix}
\mean_\ingrp^{\timeidx+1} \\
\mean_\outgrp^{\timeidx+1}
\end{pmatrix}
= (1 - \leak) \cdot
\begin{pmatrix}
\mean_\ingrp^\timeidx \\
\mean_\outgrp^\timeidx
\end{pmatrix}
+ \win \cdot
\begin{pmatrix}
\exp(-\frac{\thresh^\timeidx}{\mean_\ingrp^\timeidx}) \\
\exp(-\frac{\thresh^\timeidx}{\mean_\outgrp^\timeidx})
\end{pmatrix} \label{eq:exp_model}
\end{equation}
where the decision threshold $\thresh^\timeidx$ is selected as the numeric solution to the equation $\naccpt =
\npop_\ingrp \cdot \exp(-\frac{\thresh^\timeidx}{\mean_\ingrp^\timeidx})
+ \npop_\outgrp \cdot \exp(-\frac{\thresh^\timeidx}{\mean_\outgrp^\timeidx})$, where the parameter $\leak \in [0, 1]$
quantifies the score fruction an individual loses in each time step (\enquote{leak reate}),
and where the parameter $\win \in \R^+$ quantifies the score an individual gains for
for a positive classifier decision.
Figure~\ref{fig:systems} (left) visualizes the dynamical system.

Note the connections of our model to the real-world examples mentioned before. In credit scoring,
$\quali_\persidx^\timeidx$ would correspond to the credit score, i.e.\ the capability
of an individual to pay back a loan. We would plausibly assume that the score
increases with positive classifier decisions because individuals who get a loan have additional
financial resources at their disposal and can use those to add wealth to their group \cite{Liu2018}.
Further, we would assume a nonzero leak rate $\leak$ because individuals need to cover their
expenses which may negatively affect their capability to pay back a loan.

If we apply our model to pre-trial bail assessment, the score $\quali_\persidx^\timeidx$ would
assess the likelihood of a defendant to not commit a crime until trial.
Here, we would assume that the score decreases with negative classifier decisions because
incarcerating people from a community may cut social ties and deteriorate trust in the state,
leading to a higher crime rate \cite{ONeil2016}. This effect can be modeled by a nonzero
leak rate $\leak$ and a positive score $\win$.

Also note that our model is not necessarily realistic but shows that there exist contexts
where even perfect classifiers can exhibit stable long-term inequality. We show that context
can matter, not that every context conforms to our model.

If we analyze the equilibria of this system, we first note that
$\lim_{\mean_\ingrp^\timeidx \to 0} \mean_\ingrp^{\timeidx+1}
= \lim_{\mean_\ingrp^\timeidx \to 0} (1 - \leak) \cdot \mean_\ingrp^\timeidx + \win \cdot \exp(-\frac{\thresh^\timeidx}{\mean_\ingrp^\timeidx})
= 0$, i.e.\ $\mean_\ingrp^* = 0$ is a fix point. Further, $\lim_{\mean_\ingrp^* \to 0 }\exp(-\frac{\thresh^*}{\mean_\ingrp^*}) = 0$,
i.e.\ no person from the protected group is above the threshold at that fix point.
Accordingly, we can compute the fix point threshold $\thresh^*$ only for the non-protected group, i.e.\ $\thresh^* =
\mean_\outgrp^* \cdot \log(\frac{\npop_\outgrp}{\naccpt})$. By plugging this into the fix point
equation $\mean_\outgrp^* = (1 - \leak) \cdot \mean_\outgrp^* + \win \cdot \exp(-\frac{\thresh^*}{\nMean_\outgrp^*})$
we obtain $\mean_\outgrp^* = \frac{\win}{\leak} \cdot \frac{\naccpt}{\npop_\outgrp}$,
which yields $\mean_\outgrp^* = 2.5$ for our example in Figure~\ref{fig:systems} (left).
At this fix point, we obtain a Jacobian of Equation~\ref{eq:exp_model} which is $1 - \leak$
times the identity matrix, i.e.\ both eigenvalues have an absolute value $< 1$ for $\leak > 0$, implying
stability. In Figure~\ref{fig:systems} (left) we also see that the basin of attraction
is the entire region above the diagonal, i.e.\ whenever we start with slight inequality
in favor of the non-protected group, this inequality will get amplified.

In summary, we have shown that, for our exponential distribution model,
there are always undesirable and stable equilibria in which $\mean_\ingrp^\timeidx$
degenerates to zero and the non-protected group receives all positive outcomes.
This begs the question: Can we break this undesirable dynamic?
Indeed, we can, using demographic parity.

\section{Demographic Parity Dynamics}

Demographic parity requires equal acceptance rates across groups, i.e.\
$\exp(-\frac{\thresh_\ingrp^\timeidx}{\mean_\ingrp^\timeidx}) = \exp(-\frac{\thresh_\outgrp^\timeidx}{\mean_\outgrp^\timeidx}) = \prob$
for some acceptance rate $\prob$ and group-specific thresholds $\thresh_\ingrp^\timeidx$
and $\thresh_\outgrp^\timeidx$.
We obtain $\prob$ as solution of the threshold equation $\naccpt = \npop_\ingrp \cdot \prob + \npop_\outgrp \cdot \prob$,
i.e.\ $\prob = \frac{\naccpt}{\npop_\ingrp + \npop_\outgrp} = \frac{\naccpt}{\npop}$.
By plugging this result into our fix point equation we obtain
$\mean^* = \mean_\ingrp^* = \mean_\outgrp^* = (1 - \leak) \cdot \mean^* + \win \cdot \prob
= \frac{\win}{\leak} \cdot \frac{\naccpt}{\npop}$, which
yields $\mean^* = 5/3$ for our example in Figure~\ref{eq:exp_model} (right).
For this fix point we obtain a Jacobian of Equation~\ref{eq:exp_model} of $1 - \leak$ times
the identity matrix, implying stability.

Overall, demographic parity ensures that the mean for every group converges to
the same point, such that the thresholds $\thresh_\ingrp^\timeidx$ and
$\thresh_\outgrp^\timeidx$ become equal as well. This, in turn, implies that
selecting the top-scored people in each group corresponds to selecting the
top-scored people in the entire population, implying
a classifier that is perfectly accurate \emph{and} conforms to all notions of
fairness, including demographic parity.

\section{Conclusion}

In this contribution, we have analyzed a simple dynamic model for automatic decision making.
In particular, our model assumes that people should receive a positive classifier decision
only if their objective risk score is in the top, that the means of the score distribution
differ between the protected group and everyone else, and that positive decisions improve
the mean for the group in the next time step. This feedback loop becomes a vicious cycle
in which even a perfectly accurate classifier conforming to almost all fairness notions
leads to stable inequality. Fortunately, we can break this vicious cycle by imposing
democratic parity which instead leads to an equilibrium with perfectly accurate, equal,
and fair classification.

At present, our analysis is limited to a theoretical model assuming an exponential
distribution and a simple dynamic model. However, we note that generalizations to other
distributions are possible. Further, we note that our findings
are consistent with practical application scenarios \cite{ONeil2016} and other theoretic
studies \cite{Hu2018,Mouzannar2019}.

Overall, we conclude that our findings give reason to re-think notions of fairness in terms of
mid- and long-term outcomes and reconsider demographic parity as a helpful intervention whenever
decision making systems are embedded in vicious cycles. Otherwise, even well-intended
and well-constructed systems may stabilize and exacerbate inequality.

\bibliography{literature} 
\bibliographystyle{plainnat}

\clearpage

\begin{appendix}
\section{Generalized Setup}
\label{appendix}

In this appendix, we generalize our argument from the main paper to general probability densities.
Further, we perform the stability analysis in more detail for three common probability distributions,
namely the exponential distribution (Section~\ref{sec:exponential}), the Pareto distribution (Section~\ref{sec:pareto}), and the Gaussian distribution (Section~\ref{sec:gaussian}).

Our generalized setup is as follows. We assume a population with $\npop$ individuals, a subset of which
belong to a protected group $\group \subseteq \{1, \ldots, \npop\}$. We denote the size of the protected
group as $\npop_\ingrp = |\group|$ and the size of the non-protected group as $\npop_\outgrp = \npop - \npop_\ingrp$.
We generally assume that $0 < \npop_\ingrp < \npop_\outgrp < \npop$.

Further, we assume that every individual $\persidx \in \{1, \ldots, \npop\}$ has an objective risk score
$\quali_\persidx^\timeidx$ at time $\timeidx$, which is a real-valued random variable that is distributed
according to some density $\dens_\ingrp^\timeidx$ if $\persidx \in \group$ or according to
some density $\dens_\outgrp^\timeidx$ if $\persidx \in \neg \group$. Note that these densities may change over time
and are thus indexed with the time step $\timeidx$. Further, we assume that an individual $\persidx$ is successful,
i.e.\ $\lbl_\persidx = 1$, if and only if $\quali_\persidx^\timeidx$ is among the top $\naccpt$ scores at time
$\timeidx$. We also assume that $0 < \naccpt \ll \npop_\ingrp$, i.e.\ an acceptance is rare.

Under these assumptions, the following classifier $\fun^\timeidx : \{1, \ldots, \npop\} \to \{0, 1\}$ is
per construction perfectly accurate.
\begin{align*}
\fun^\timeidx(\persidx) &=
\begin{cases}
1 & \text{if } \scr^\timeidx(\persidx) \geq \thresh^\timeidx \\
0 & \text{otherwise}
\end{cases} & \text{where} \\
\scr^\timeidx(\persidx) &= \quali_\persidx^\timeidx \\
\thresh^\timeidx &\text{ s.t.} \quad |\{ \persidx | \quali_\persidx^\timeidx \geq \thresh^\timeidx \}| = \naccpt
\end{align*}
In other words, we predict success for individual $\persidx$ at time $\timeidx$ by first assigning the
risk score $\quali_\persidx^\timeidx$ and then applying a threshold that only leaves the top $\naccpt$
people, i.e.\ exactly those who actually will be successful.

Note that this classifier is not only perfectly accurate but also conforms to equalized odds because we
do not misclassify anyone, such that the rate of misclassifications is equal among all groups \cite{Hardt2016}.
Further, the scoring function is calibrated, the classifier is as accurate as possible, and it applies the
same threshold for everyone, such that the fairness constraints of \citet{Corbett2018} are fulfilled.
Finally, the classifier fulfills notions of due process because it only accesses the objective risk score
and makes no use of any other features of the individual \cite{ONeil2016,Dwork2012,Grgic2016,Kilbertus2017,Kusner2017}.

Next, we consider the probability of a person inside our outside the protected group to be classified as
successful at time $\timeidx$, which is both equivalent to the probability of being successful at time $\timeidx$
and to the probability of having a score above or equal to the threshold $\thresh^\timeidx$
at time $\timeidx$. We denote this probability as $\prob_\ingrp^\timeidx$ for the protected group and
$\prob_\outgrp^\timeidx$ otherwise. These probabilities are given as follows.
\begin{align*}
\prob_\ingrp^\timeidx &= \int_{\thresh^\timeidx}^\infty \dens_\ingrp^\timeidx(\quali) d\quali \\
\prob_\ingrp^\timeidx &= \int_{\thresh^\timeidx}^\infty \dens_\ingrp^\timeidx(\quali) d\quali
\end{align*}
Given these probabilities, we can estimate the number of people inside and outside the protected group
who will be successful at time $\timeidx$. For each individual, success at time $\timeidx$ is an independent
Bernoulli trial with success probability $\prob_\ingrp^\timeidx$ or $\prob_\outgrp^\timeidx$ respectively.
Accordingly, the sums $\sum_{\persidx \in \group} \fun(\persidx)$ and $\sum_{\persidx \in \neg \group}
\fun(\persidx)$ are binomially distributed random variables with means $\npop_\ingrp \cdot \prob_\ingrp^\timeidx$
as well as $\npop_\outgrp \cdot \prob_\outgrp^\timeidx$ and variances
$\npop_\ingrp \cdot \prob_\ingrp^\timeidx \cdot (1 - \prob_\ingrp^\timeidx)$ as well as
$\npop_\outgrp \cdot \prob_\outgrp^\timeidx \cdot (1 - \prob_\outgrp^\timeidx)$.
Note that the probabilities $\prob_\ingrp^\timeidx$ with $\prob_\outgrp^\timeidx$ decrease for lower $\naccpt$.
Therefore, our assumption $\naccpt \ll \npop_\ingrp < \npop_\outgrp$ implies that the variance is close to zero
and thus the mean is a precise estimate of the actual number of successful people.

\begin{figure}
\begin{center}
\begin{tikzpicture}
\begin{axis}[xlabel={$\quali$}, ylabel={no.\ people},
xmin=0,xmax=15,ymin=0,ymax=35,
extra x ticks={5.385},
extra x tick style={grid=major,aluminium4},
extra x tick labels={$\thresh^\timeidx$},
legend cell align=left,
legend pos=outer north east,]
% out group density
\addplot[draw=orange3, name path=dens_outgrp, domain=0:15] {exp(-x/3)/3*100};
\addlegendentry{$\dens_\outgrp^\timeidx(\quali) \cdot \npop_\outgrp$}
% in group density
\addplot[draw=skyblue3, name path=dens_ingrp, domain=0:15] {exp(-x/2)/2*50};
\addlegendentry{$\dens_\ingrp^\timeidx(\quali) \cdot \npop_\ingrp$}
% fill between in group and out group
\addplot[orange1] fill between[of=dens_ingrp and dens_outgrp, soft clip={domain=5.385:15}];
\addlegendentry{$\prob_\outgrp^\timeidx \cdot \npop_\outgrp = \int_{\thresh^\timeidx}^\infty \dens_\outgrp^\timeidx(\quali) d\quali \cdot \npop_\outgrp$}
% fill between in group and bottom
\path[name path=bot]
(\pgfkeysvalueof{/pgfplots/xmin},0) --
(\pgfkeysvalueof{/pgfplots/xmax},0);
\addplot[skyblue1] fill between[of=dens_ingrp and bot, soft clip={domain=5.385:15}];
\addlegendentry{$\prob_\ingrp^\timeidx \cdot \npop_\ingrp = \int_{\thresh^\timeidx}^\infty \dens_\ingrp^\timeidx(\quali) d\quali \cdot \npop_\ingrp$}
% acceptance limits
\addplot[black, domain=0:15] {20};
\node[above right] at (axis cs:5.5,20) {$\naccpt = \npop_\ingrp \cdot \prob_\ingrp^\timeidx
	+ \npop_\outgrp \cdot \prob_\outgrp^\timeidx$};
\end{axis}
\end{tikzpicture}
\end{center}
\caption{An illustration of Equation~\ref{eq:threshold}
for $\naccpt = 20$, $\npop_\ingrp = 50$, $\npop_\outgrp = 100$, $\mean_\ingrp^\timeidx = 2$, and $\mean_\outgrp^\timeidx = 3$.
The threshold $\thresh^\timeidx$ is selected on the x-axis
such that exactly the amount of probability mass from both $\dens_\ingrp^\timeidx$ and $\dens_\outgrp^\timeidx$
lies on the right side of $\thresh^\timeidx$ so that $\npop_\ingrp \cdot \prob_\ingrp^\timeidx$ and
$\npop_\outgrp \cdot \prob_\outgrp^\timeidx$ add up to $\npop$.}
\label{fig:threshold}
\end{figure}

Using our mean estimate for the number of successful people in each group, the threshold $\thresh^\timeidx$
can be estimated using the approximate equation
\begin{equation}
\naccpt = \npop_\ingrp \cdot \prob_\ingrp^\timeidx + \npop_\outgrp \cdot \prob_\outgrp^\timeidx. \label{eq:threshold}
\end{equation}
Figuratively speaking, we slide $\thresh^\timeidx$ from right to left
along the score axis until we have collected enough probability mass from both $\dens_\ingrp^\timeidx$
and $\dens_\outgrp^\timeidx$ such that exactly $\naccpt$ people are expected to have a score above
the threshold (also refer to Figure~\ref{fig:threshold}).

\subsection{Dynamical Model}

Finally, we model the dynamics of our system. Our central modeling decision is that the means of the
densities $\dens_\ingrp^\timeidx$ and $\dens_\outgrp^\timeidx$ shift over time. More precisely, we denote the
means at time $\timeidx$ as $\mean_\ingrp^\timeidx$ and $\mean_\outgrp^\timeidx$, and assume the following
dynamical system equation.
\begin{equation}
\begin{pmatrix}
\mean_\ingrp^{\timeidx+1} \\
\mean_\outgrp^{\timeidx+1}
\end{pmatrix}
= (1 - \leak) \cdot
\begin{pmatrix}
\mean_\ingrp^\timeidx \\
\mean_\outgrp^\timeidx
\end{pmatrix}
+ \win \cdot
\begin{pmatrix}
\prob_\ingrp^\timeidx \\
\prob_\outgrp^\timeidx
\end{pmatrix} \label{eq:general_model}
\end{equation}
where $\leak \in [0, 1]$ is a hyper-parameter describing how much score an individual loses in each time step (\enquote{leak reate}),
and where $\win \in \R^+$ is a hyper-parameter describing how much score an individual gains for
for each positive classifier decision. By averaging these scores inside and outside the
protected group, we obtain Equation~\ref{eq:general_model}. Note that the dynamics of
$\mean_\ingrp^\timeidx$ and $\mean_\outgrp^\timeidx$ are implicitly coupled due to the shared threshold
$\thresh^\timeidx$.

Now, let us analyze the dynamic behavior of this system. In particular, we consider the score
gap between the protected group and everyone else, which we denote as $\gap^\timeidx := \mean_\outgrp^\timeidx - \mean_\ingrp^\timeidx$.
This score gap grows at time step $\timeidx$ exactly if
\begin{align}
&&\gap^{\timeidx+1} - \gap^\timeidx &> 0 \notag \\
\iff &&\mean_\outgrp^{\timeidx+1} - \mean_\ingrp^{\timeidx+1} - \mean_\outgrp^\timeidx + \mean_\ingrp^\timeidx &> 0 \notag \\
\iff &&(1 - \leak) \cdot \mean_\outgrp^\timeidx + \win \cdot \prob_\outgrp^\timeidx - (1 - \leak) \cdot \mean_\ingrp^\timeidx - \win \cdot \prob_\ingrp^\timeidx - \mean_\outgrp^\timeidx + \mean_\ingrp^\timeidx &> 0 \notag \\
\iff &&\win \cdot \big(\prob_\outgrp^\timeidx - \prob_\ingrp^\timeidx) &> \leak \cdot \gap^\timeidx \label{eq:growth}.
\end{align}
Conversely, the score gap shrinks at time step at $\timeidx$ exactly if
\begin{equation}
\win \cdot \big(\prob_\outgrp^\timeidx - \prob_\ingrp^\timeidx) < \leak \cdot \gap^\timeidx \label{eq:shrink}.
\end{equation}

From these equations, we can infer that the absolute value of the score gap will grow if $\leak$ is sufficiently small,
$\win$ is sufficiently large, and the probability distribution emphasizes gaps in the mean,
i.e.\ small differences between means imply larger differences in probability
mass on the margins. This covers a broad range of distributions where probability mass is concentrated
around the mean.

\subsection{Equilibria}

Now, let us analyze the equilibria of our dynamical system. First, let us consider undesirable
equilibria with a nonzero score gap. In particular, let us assume that
the protected group is entirely unsuccessful, i.e.\ $\prob_\ingrp^* = 0$. Then, by virtue of Equation~\ref{eq:threshold},
we obtain $\prob_\outgrp^* = \frac{\naccpt}{\npop_\outgrp}$. The fix point equation yields:
\begin{align*}
\begin{pmatrix}
\mean_\ingrp^*\\
\mean_\outgrp^*
\end{pmatrix}
= (1 - \leak) \cdot
\begin{pmatrix}
\mean_\ingrp^* \\
\mean_\outgrp^*
\end{pmatrix}
+ \win \cdot
\begin{pmatrix}
0 \\
\frac{\naccpt}{\npop_\outgrp}
\end{pmatrix}
\quad \iff
\begin{pmatrix}
\mean_\ingrp^*\\
\mean_\outgrp^*
\end{pmatrix}
= 
\begin{pmatrix}
0 \\
\frac{\win}{\leak} \cdot \frac{\naccpt}{\npop_\outgrp}
\end{pmatrix}
\end{align*}
In other words, we achieve an equilibrium if the protected group has a mean score of zero and no success
whereas everybody else has a mean score of $\frac{\win}{\leak} \cdot \frac{\naccpt}{\npop_\outgrp}$ and
success probability $\frac{\naccpt}{\npop_\outgrp}$. For many probability distributions, we can assume
that probabilities remain unaffected by small deviations in the score gap. Therefore, we can demonstrate
stability as follows.

First, assume that the score gap is slightly larger compared to the equilibrium. In this case, we obtain:
\begin{equation*}
\leak \cdot \gap^\timeidx > \leak \cdot( \mean_\outgrp^* - \mean_\ingrp^* )
= \leak \cdot \frac{\win}{\leak} \cdot \frac{\naccpt}{\npop_\outgrp}
= \win \cdot \frac{\naccpt}{\npop_\outgrp}
= \win \cdot (\prob^*_\outgrp - \prob^*_\ingrp) \approx \win \cdot(\prob^\timeidx_\outgrp - \prob^\timeidx_\ingrp)
\end{equation*}
which in turn implies by virtue of Equation~\ref{eq:shrink} that the score gap will shrink again.
Second, assume that the score gap is slightly smaller compared to the equilibrium. In this case, we obtain:
\begin{equation*}
\leak \cdot \gap^\timeidx < \leak \cdot( \mean_\outgrp^* - \mean_\ingrp^* )
= \leak \cdot \frac{\win}{\leak} \cdot \frac{\naccpt}{\npop_\outgrp}
= \win \cdot \frac{\naccpt}{\npop_\outgrp}
= \win \cdot (\prob^*_\outgrp - \prob^*_\ingrp) \approx \win \cdot(\prob^\timeidx_\outgrp - \prob^\timeidx_\ingrp)
\end{equation*}
which in turn implies by virtue of Equation~\ref{eq:growth} that the score gap grows again.
Overall, the dynamical system counteracts small deviations, implying stability.

Next, we consider the desirable case, where success probabilities become equal across groups,
i.e.\ $\prob_\ingrp^* = \prob_\outgrp^* = \prob^*$.
Then, Equation~\ref{eq:threshold} yields $\naccpt = \prob^* \cdot \npop_\ingrp + \prob^* \cdot \npop_\outgrp$,
which implies $\prob^* = \frac{\naccpt}{\npop}$.
Accordingly, the fix point equation yields:
\begin{equation*}
\begin{pmatrix}
\mean_\ingrp^* \\
\mean_\outgrp^* 
\end{pmatrix}
= (1 - \leak) \cdot
\begin{pmatrix}
\mean_\ingrp^* \\
\mean_\outgrp^*
\end{pmatrix}
+ \win \cdot
\begin{pmatrix}
\prob^* \\
\prob^*
\end{pmatrix}
\quad
\iff
\begin{pmatrix}
\mean_\ingrp^* \\
\mean_\outgrp^* 
\end{pmatrix}
= \begin{pmatrix}
\frac{\win}{\leak} \cdot \frac{\naccpt}{\npop} \\
\frac{\win}{\leak} \cdot \frac{\naccpt}{\npop}
\end{pmatrix}
\end{equation*}
In other words, the mean for both groups has the same value $\mean_\ingrp^* = \mean_\outgrp^* = \frac{\win}{\leak} \cdot \frac{\naccpt}{\npop}$.
Unfortunately, this point is not stable in general, because a slight nonzero
score gap will, for many probability distributions, result in emphasized gaps in success probability, which
in turn may fulfill Equation~\ref{eq:growth} or~\ref{eq:shrink}, leading to even more pronounced score gaps
(as we will show in Sections~\ref{sec:exponential}, \ref{sec:pareto}, and \ref{sec:gaussian}).
To counteract this potential instability, we can employ demographic parity.

\subsection{Demographic Parity}

Demographic parity requires that, at any time step $\timeidx$, $\prob_\ingrp^\timeidx = \prob_\outgrp^\timeidx$,
which we can re-write to $\prob_\outgrp^\timeidx - \prob_\ingrp^\timeidx = 0$. This, in turn, implies that
Equation~\ref{eq:growth} is always fulfilled if the score gap is negative and Equation~\ref{eq:shrink} is always
fulfilled if the score gap is positive. In other words, demographic parity ensures that only states with zero
score gaps can be equilibria and that these equilibria are always stable. In case there is a one-to-one map
from means to success probabilities (as is the case in most common probability distributions), this in turn
implies that there is a unique, stable equilibrium where the score gap and the probability gap are both zero,
i.e.\ $\prob_\ingrp^* = \prob_\outgrp^* = \frac{\naccpt}{\npop}$ and $\mean_\ingrp^* = \mean_\outgrp^* =
\frac{\win}{\leak} \cdot \frac{\naccpt}{\npop}$.

We can also confirm the stability finding using classic stability theory. If $\prob_\ingrp^\timeidx =
\prob_\outgrp^\timeidx$, then Equation~\ref{eq:threshold} implies that $\prob_\ingrp^\timeidx =
\prob_\outgrp^\timeidx = \frac{\naccpt}{\npop}$ for all time steps. Therefore, we obtain the following
Jacobian for Equation~\ref{eq:general_model}
\begin{equation*}
\jac(\mean_\ingrp^\timeidx, \mean_\outgrp^\timeidx) =
\begin{pmatrix}
\deriv{\mean_\ingrp^\timeidx} \big[(1 - \leak) \cdot \mean_\ingrp^\timeidx + \win \cdot \frac{\naccpt}{\npop}\big] &
\deriv{\mean_\outgrp^\timeidx} \big[(1 - \leak) \cdot \mean_\ingrp^\timeidx + \win \cdot \frac{\naccpt}{\npop}\big] \\
\deriv{\mean_\ingrp^\timeidx} \big[(1 - \leak) \cdot \mean_\outgrp^\timeidx + \win \cdot \frac{\naccpt}{\npop}\big] &
\deriv{\mean_\outgrp^\timeidx} \big[(1 - \leak) \cdot \mean_\outgrp^\timeidx + \win \cdot \frac{\naccpt}{\npop}\big]
\end{pmatrix}
=
\begin{pmatrix}
1 - \leak & 0 \\
0 & 1 - \leak
\end{pmatrix}
\end{equation*}
The two eigenvalues of this matrix are both $1 - \leak$. Therefore, for a nonzero leak rate, the
absolute value of these eigenvalues is smaller than one and therefore stability theory implies that the
equilibrium is stable. Note that this result applies independent of the probability distribution
in question and relies only on a one-to-one mapping between means and success probabilities.

In the following sections, we consider the stability findings from the previous section in more detail
for three specific distributions, namely the exponential, the Pareto, and the Gaussian distribution.
 
\subsection{Stability for the Exponential Distribution}
\label{sec:exponential}

In the exponential distribution model, we assume that the densities $\dens_\ingrp^\timeidx$ and $\dens_\outgrp^\timeidx$
have the following form (also refer to Figure~\ref{fig:threshold}).
\begin{align*}
\dens_\ingrp^\timeidx(\quali) &= \frac{1}{\mean_\ingrp^\timeidx} \cdot \exp\big(-\frac{\quali}{\mean_\ingrp^\timeidx}\big) \\
\dens_\outgrp^\timeidx(\quali) &= \frac{1}{\mean_\outgrp^\timeidx} \cdot \exp\big(-\frac{\quali}{\mean_\outgrp^\timeidx}\big)
\end{align*}
Note that the densities are fully parameterized by the respective mean, which makes the exponential
distribution a straightforward object of study. For these densities, the probabilities $\prob_\ingrp^\timeidx$
and $\prob_\outgrp^\timeidx$ take the following form.
\begin{align*}
\prob_\ingrp^\timeidx &= \int_{\thresh^\timeidx}^\infty \dens_\ingrp^\timeidx(\quali) d\quali = \exp\big(-\frac{\thresh^\timeidx}{\mean_\ingrp^\timeidx}\big)\\
\prob_\outgrp^\timeidx &= \int_{\thresh^\timeidx}^\infty \dens_\outgrp^\timeidx(\quali) d\quali = \exp\big(-\frac{\thresh^\timeidx}{\mean_\outgrp^\timeidx}\big)
\end{align*}

Now, let us consider the undesirable equilibrium case, where $\mean_\ingrp^* = 0$ and $\mean_\outgrp^*
= \frac{\win}{\leak} \cdot \frac{\naccpt}{\npop_\outgrp}$. First, note that the threshold $\thresh^\timeidx$
in this condition is lower-bounded due to Equation~\ref{eq:threshold}. In particular, we obtain the following lower bound.
\begin{align*}
&&\exp\big(-\frac{\thresh^\timeidx}{\mean_\outgrp^*}\big)
= \exp\big(-\thresh^\timeidx \cdot \frac{\leak}{\win} \cdot \frac{\npop_\outgrp}{\naccpt}\big)
&\leq \frac{\naccpt}{\npop_\outgrp} \\
\iff && -\thresh^\timeidx \cdot \frac{\leak}{\win} \cdot \frac{\npop_\outgrp}{\naccpt} &\leq \log\big(\frac{\naccpt}{\npop_\outgrp}\big) \\
\iff && \thresh^\timeidx &\geq - \frac{\win}{\leak} \cdot \frac{\naccpt}{\npop_\outgrp} \log\big(\frac{\naccpt}{\npop_\outgrp}\big)
\end{align*}
Note that this term is strictly larger than zero as $\naccpt \ll \npop_\outgrp$, which implies
that $\log\big(\frac{\naccpt}{\npop_\outgrp}\big) < 0$.
Accordingly, if $\mean_\ingrp^\timeidx$ is sufficiently small, i.e.\ $\mean_\ingrp^\timeidx \ll
- \frac{\win}{\leak} \cdot \frac{\naccpt}{\npop_\outgrp} \log\big(\frac{\naccpt}{\npop_\outgrp}\big)$,
then $\prob_\ingrp^\timeidx = \exp\big(-\frac{\thresh^\timeidx}{\mean_\ingrp^\timeidx}\big) \approx 0$.
Also note that small changes in $\mean_\ingrp^\timeidx$ will not change the probability $\prob_\ingrp^\timeidx$
in this case. Therefore, for sufficiently small $\mean_\ingrp^\timeidx$, we obtain approximatively constant
probabilities $\prob_\ingrp^\timeidx \approx 0$ and $\prob_\outgrp^\timeidx \approx \frac{\naccpt}{\npop_\outgrp}$.
Accordingly, the Jacobian matrix of Equation~\ref{eq:general_model} for sufficiently small
$\mean_\ingrp^\timeidx$ is given as follows.
\begin{equation*}
\jac(\mean_\ingrp^\timeidx, \mean_\outgrp^\timeidx) =
\begin{pmatrix}
\deriv{\mean_\ingrp^\timeidx} \big[(1 - \leak) \cdot \mean_\ingrp^\timeidx + \win \cdot 0\big] &
\deriv{\mean_\outgrp^\timeidx} \big[(1 - \leak) \cdot \mean_\ingrp^\timeidx + \win \cdot 0\big] \\
\deriv{\mean_\ingrp^\timeidx} \big[(1 - \leak) \cdot \mean_\outgrp^\timeidx + \win \cdot \frac{\naccpt}{\npop_\outgrp}\big] &
\deriv{\mean_\outgrp^\timeidx} \big[(1 - \leak) \cdot \mean_\outgrp^\timeidx + \win \cdot \frac{\naccpt}{\npop_\outgrp}\big]
\end{pmatrix}
=
\begin{pmatrix}
1 - \leak & 0 \\
0 & 1 - \leak
\end{pmatrix}
\end{equation*}
The two eigenvalues of this matrix are both $1 - \leak$. Therefore, for a nonzero leak rate, the
absolute value of these eigenvalues is smaller than one and therefore stability theory implies that the
equilibrium is stable.

Next, let us consider the desirable equilibrium case, where $\mean_\ingrp^* = \mean_\outgrp^* =
\frac{\win}{\leak} \cdot \frac{\naccpt}{\npop}$ and $\prob_\ingrp^* = \prob_\outgrp^* = \frac{\naccpt}{\npop}$.
In this equilibrium condition, we can obtain the threshold $\thresh^*$ as follows.
\begin{align*}
&&\frac{\naccpt}{\npop} &= \prob_\ingrp^* = \exp\big(-\frac{\thresh^*}{\mean_\ingrp^*}\big) \\
\iff && \log\big(\frac{\naccpt}{\npop}\big) &= - \frac{\thresh^*}{\mean_\ingrp^*} \\
\iff && \thresh^* &= - \mean_\ingrp^* \cdot \log\big(\frac{\naccpt}{\npop}\big)
\end{align*}
Now, let us consider small deviations of $\mean_\ingrp^\timeidx$ and $\mean_\outgrp^\timeidx$
such that $\thresh^\timeidx$ stays equal to $\thresh^*$. Such deviations are possible since we can
let both means deviate in opposite directions such that the condition holds. In that case,
we obtain the following Jacobian of our model in Equation~\ref{eq:general_model}.
\begin{align*}
\jac(\mean_\ingrp^\timeidx, \mean_\outgrp^\timeidx) &=
\begin{pmatrix}
\deriv{\mean_\ingrp^\timeidx} \big[(1 - \leak) \cdot \mean_\ingrp^\timeidx + \win \cdot \exp\big(-\frac{\thresh^*}{\mean_\ingrp^\timeidx}\big)\big] &
\deriv{\mean_\outgrp^\timeidx} \big[(1 - \leak) \cdot \mean_\ingrp^\timeidx + \win \cdot \exp\big(-\frac{\thresh^*}{\mean_\ingrp^\timeidx}\big)\big] \\
\deriv{\mean_\ingrp^\timeidx} \big[(1 - \leak) \cdot \mean_\outgrp^\timeidx + \win \cdot \exp\big(-\frac{\thresh^*}{\mean_\outgrp^\timeidx}\big)\big] &
\deriv{\mean_\outgrp^\timeidx} \big[(1 - \leak) \cdot \mean_\outgrp^\timeidx + \win \cdot \exp\big(-\frac{\thresh^*}{\mean_\outgrp^\timeidx}\big)\big]
\end{pmatrix} \\
&=
\begin{pmatrix}
1 - \leak + \win \cdot \exp\big(-\frac{\thresh^*}{\mean_\ingrp^\timeidx}\big) \cdot \frac{\thresh^*}{(\mean_\ingrp^\timeidx)^2} & 0 \\
0 & 1 - \leak + \win \cdot \exp\big(-\frac{\thresh^*}{\mean_\outgrp^\timeidx}\big) \cdot \frac{\thresh^*}{(\mean_\outgrp^\timeidx)^2}
\end{pmatrix}
\end{align*}
Accordingly, the Jacobian at our equilibrium is given as follows.
\begin{align*}
\jac(\mean_\ingrp^*, \mean_\outgrp^*)
&= \begin{pmatrix}
1 - \leak + \win \cdot \exp\big(-\frac{\thresh^*}{\mean_\ingrp^*}\big) \cdot \frac{\thresh^*}{(\mean_\ingrp^*)^2} & 0\\
0 & 1 - \leak + \win \cdot \exp\big(-\frac{\thresh^*}{\mean_\outgrp^*}\big) \cdot \frac{\thresh^*}{(\mean_\outgrp^*)^2}\\
\end{pmatrix} \\
&= \begin{pmatrix}
1 - \leak - \win \cdot \exp\big(\log\big(\frac{\naccpt}{\npop}\big)\big) \cdot \frac{\leak}{\win} \cdot \frac{\npop}{\naccpt} \cdot \log\big(\frac{\naccpt}{\npop}\big) & 0\\
0 & 1 - \leak - \win \cdot \exp\big(\log\big(\frac{\naccpt}{\npop}\big)\big) \cdot \frac{\leak}{\win} \cdot \frac{\npop}{\naccpt} \cdot \log\big(\frac{\naccpt}{\npop}\big)\\
\end{pmatrix} \\
&= \begin{pmatrix}
1 - \leak - \leak \cdot \log\big(\frac{\naccpt}{\npop}\big) & 0\\
0 & 1 - \leak - \leak \cdot \log\big(\frac{\naccpt}{\npop}\big)\\
\end{pmatrix}
\end{align*}
The two eigenvalues of this Jacobian are both $1 - \leak - \leak \cdot \log\big(\frac{\naccpt}{\npop}\big)$.
Accordingly, our equilibrium is unstable if $|1 - \leak - \leak \cdot \log\big(\frac{\naccpt}{\npop}\big)| > 1$.
Given that $\naccpt \ll \npop$ and $\leak \in [0, 1]$, $1 - \leak - \leak \cdot \log\big(\frac{\naccpt}{\npop}\big)$
is larger than $0$. Therefore, we can re-write the instability condition as follows.
\begin{align*}
&&1 - \leak - \leak \cdot \log\big(\frac{\naccpt}{\npop}\big) &> 1 \\
\iff && 1 + \log\big(\frac{\naccpt}{\npop}\big) &< 0 \\
\iff && \frac{\naccpt}{\npop} &< \frac{1}{e}
\end{align*}
In other words, if $\npop$ is at least $e$ times larger than $\naccpt$, the equilibrium is unstable.
From $\naccpt \ll \npop_\ingrp < \frac{1}{2} \npop$, we can conclude that this is the case.

In summary, we have demonstrated that the exponential distribution yields attractive equilibria in
undesirable positions, whereas the desirable equilibria are unstable for a wide range of conditions.

\subsection{Stability for the Pareto Distribution}
\label{sec:pareto}

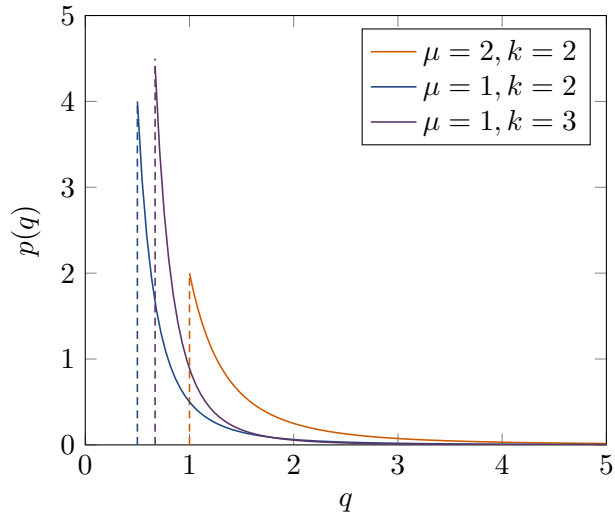
\begin{figure}
\begin{center}
\begin{tikzpicture}
\begin{axis}[xlabel={$\quali$}, ylabel={$\dens(\quali)$},
xmin=0,xmax=5,ymin=0,ymax=5,
legend cell align=left,
legend pos=north east,]
\addplot[draw=orange3, semithick, name path=dens_outgrp, domain=1:5, samples=100] {0.5 * 2^2 / x^3};
\draw[orange3, semithick, densely dashed] (axis cs:1,0) -- (axis cs:1,2);
\addlegendentry{$\mean = 2, \parrate = 2$}
\addplot[draw=skyblue3, semithick, name path=dens_ingrp, domain=0.5:5, samples=100] {0.5 / x^3};
\draw[skyblue3, semithick, densely dashed] (axis cs:0.5,0) -- (axis cs:0.5,4);
\addlegendentry{$\mean = 1, \parrate = 2$}
\addplot[draw=plum3, semithick, name path=dens_ingrp, domain=0.67:5, samples=100] {(2^3/3^2) / x^4};
\draw[plum3, semithick, densely dashed] (axis cs:0.67,0) -- (axis cs:0.67,4.5);
\addlegendentry{$\mean = 1, \parrate = 3$}
\end{axis}
\end{tikzpicture}
\end{center}
\caption{Three different Pareto densities with means $\mean$ and rate parameters $\parrate$ as specified in the legend.}
\label{fig:pareto}
\end{figure}

In the Pareto distribution model, we assume that the densities
$\dens_\ingrp^\timeidx$ and $\dens_\outgrp^\timeidx$
have the following form (also refer to Figure~\ref{fig:pareto}).
\begin{align*}
\dens_\ingrp^\timeidx(\quali) &= \frac{(\parrate-1)^\parrate}{\parrate^{\parrate-1}} \cdot \frac{(\mean_\ingrp^\timeidx)^\parrate}{\quali^{\parrate+1}} \\
\dens_\outgrp^\timeidx(\quali) &= \frac{(\parrate-1)^\parrate}{\parrate^{\parrate-1}} \cdot \frac{(\mean_\outgrp^\timeidx)^\parrate}{\quali^{\parrate+1}}
\end{align*}
where $\parrate > 1$ controls the slope of the
distribution and how much probability mass is concentrated around the mean.
Note that the Pareto distribution is only defined on the interval $[\frac{\parrate-1}{\parrate} \cdot \mean, \infty)$
and that the variance is infinite for $\parrate \leq 2$.

From these densities we obtain the following success probabilities $\prob_\ingrp^\timeidx$ and
$\prob_\outgrp^\timeidx$.
\begin{align*}
\prob_\ingrp^\timeidx &= \int_{\thresh^\timeidx}^\infty \dens_\ingrp^\timeidx(\quali) d\quali =
\Big(\frac{\parrate-1}{\parrate}\Big)^\parrate \cdot \Big(\frac{\mean_\ingrp^\timeidx}{\thresh^\timeidx} \Big)^\parrate\\
\prob_\outgrp^\timeidx &= \int_{\thresh^\timeidx}^\infty \dens_\outgrp^\timeidx(\quali) d\quali = \Big(\frac{\parrate-1}{\parrate}\Big)^\parrate \cdot \Big(\frac{\mean_\outgrp^\timeidx}{\thresh^\timeidx} \Big)^\parrate
\end{align*}
Plugging these results into Equation~\ref{eq:threshold}, we obtain a closed-form expression for the
threshold as follows.
\begin{align*}
&&\naccpt &= \npop_\ingrp \cdot \Big(\frac{\parrate-1}{\parrate}\Big)^\parrate \cdot \Big(\frac{\mean_\ingrp^\timeidx}{\thresh^\timeidx} \Big)^\parrate +
\npop_\outgrp \cdot \Big(\frac{\parrate-1}{\parrate}\Big)^\parrate \cdot \Big(\frac{\mean_\outgrp^\timeidx}{\thresh^\timeidx} \Big)^\parrate \\
\iff && \naccpt \cdot (\thresh^\timeidx)^\parrate &= \Big(\frac{\parrate-1}{\parrate}\Big)^\parrate
\cdot \big[\npop_\ingrp \cdot (\mean_\ingrp^\timeidx)^\parrate + \npop_\outgrp \cdot (\mean_\outgrp^\timeidx)^\parrate \big] \\
\iff && \thresh^\timeidx &= \frac{\parrate-1}{\parrate} \cdot \sqrt[\uproot{6}\parrate]{
\frac{1}{\naccpt} \big[\npop_\ingrp \cdot (\mean_\ingrp^\timeidx)^\parrate + \npop_\outgrp \cdot (\mean_\outgrp^\timeidx)^\parrate \big]}
\end{align*}
Next, we plug this solution back into our probability expressions and obtain:
\begin{align*}
\prob_\ingrp^\timeidx &= \frac{\naccpt \cdot (\mean_\ingrp^\timeidx)^\parrate}{\npop_\ingrp \cdot (\mean_\ingrp^\timeidx)^\parrate + \npop_\outgrp \cdot (\mean_\outgrp^\timeidx)^\parrate} \\
\prob_\outgrp^\timeidx &= \frac{\naccpt \cdot (\mean_\outgrp^\timeidx)^\parrate}{\npop_\ingrp \cdot (\mean_\ingrp^\timeidx)^\parrate + \npop_\outgrp \cdot (\mean_\outgrp^\timeidx)^\parrate}
\end{align*}
Accordingly, we obtain the following Jacobian.
\begin{align*}
&\jac(\mean_\ingrp^\timeidx, \mean_\outgrp^\timeidx) \\
=&\begin{pmatrix}
\deriv{\mean_\ingrp^\timeidx} \Big[(1 - \leak) \cdot \mean_\ingrp^\timeidx + \win \cdot \frac{\naccpt \cdot (\mean_\ingrp^\timeidx)^\parrate}{\npop_\ingrp \cdot (\mean_\ingrp^\timeidx)^\parrate + \npop_\outgrp \cdot (\mean_\outgrp^\timeidx)^\parrate}\Big] &
\deriv{\mean_\outgrp^\timeidx} \Big[(1 - \leak) \cdot \mean_\ingrp^\timeidx + \win \cdot \frac{\naccpt \cdot (\mean_\ingrp^\timeidx)^\parrate}{\npop_\ingrp \cdot (\mean_\ingrp^\timeidx)^\parrate + \npop_\outgrp \cdot (\mean_\outgrp^\timeidx)^\parrate}\Big] \\
\deriv{\mean_\ingrp^\timeidx} \Big[(1 - \leak) \cdot \mean_\outgrp^\timeidx + \win \frac{\naccpt \cdot (\mean_\outgrp^\timeidx)^\parrate}{\npop_\ingrp \cdot (\mean_\ingrp^\timeidx)^\parrate + \npop_\outgrp \cdot (\mean_\outgrp^\timeidx)^\parrate}\Big] &
\deriv{\mean_\outgrp^\timeidx} \Big[(1 - \leak) \cdot \mean_\outgrp^\timeidx + \win \cdot \frac{\naccpt \cdot (\mean_\outgrp^\timeidx)^\parrate}{\npop_\ingrp \cdot (\mean_\ingrp^\timeidx)^\parrate + \npop_\outgrp \cdot (\mean_\outgrp^\timeidx)^\parrate}\Big]
\end{pmatrix} \\
=&\begin{pmatrix}
1 - \leak + \win \cdot \parrate \cdot \naccpt \cdot (\mean_\ingrp^\timeidx)^{\parrate-1} \cdot \frac{\npop_\outgrp \cdot (\mean_\outgrp^\timeidx)^\parrate}{(\npop_\ingrp \cdot (\mean_\ingrp^\timeidx)^\parrate + \npop_\outgrp \cdot (\mean_\outgrp^\timeidx)^\parrate)^2}
& -\win \cdot \frac{\naccpt \cdot (\mean_\ingrp^\timeidx)^\parrate}{(\npop_\ingrp \cdot (\mean_\ingrp^\timeidx)^\parrate + \npop_\outgrp \cdot (\mean_\outgrp^\timeidx)^\parrate)^2} \cdot \npop_\outgrp \cdot \parrate \cdot (\mean_\outgrp^\timeidx)^{\parrate-1} \\
-\win \cdot \frac{\naccpt \cdot (\mean_\outgrp^\timeidx)^\parrate}{(\npop_\ingrp \cdot (\mean_\ingrp^\timeidx)^\parrate + \npop_\outgrp \cdot (\mean_\outgrp^\timeidx)^\parrate)^2} \cdot \npop_\ingrp \cdot \parrate \cdot (\mean_\ingrp^\timeidx)^{\parrate-1}
& 1 - \leak + \win \cdot \parrate \cdot \naccpt \cdot (\mean_\outgrp^\timeidx)^{\parrate-1} \cdot \frac{\npop_\ingrp \cdot (\mean_\ingrp^\timeidx)^\parrate}{(\npop_\ingrp \cdot (\mean_\ingrp^\timeidx)^\parrate + \npop_\outgrp \cdot (\mean_\outgrp^\timeidx)^\parrate)^2}
\end{pmatrix} \\
=&\begin{pmatrix}
1 - \leak + \frac{\win \cdot \parrate \cdot (\mean_\ingrp^\timeidx)^{\parrate-1}}{\npop_\ingrp \cdot (\mean_\ingrp^\timeidx)^\parrate + \npop_\outgrp \cdot (\mean_\outgrp^\timeidx)^\parrate} \cdot \prob_\outgrp^\timeidx
& -\frac{\win \cdot \npop_\outgrp \cdot \parrate \cdot (\mean_\outgrp^\timeidx)^{\parrate-1}}{\npop_\ingrp \cdot (\mean_\ingrp^\timeidx)^\parrate + \npop_\outgrp \cdot (\mean_\outgrp^\timeidx)^\parrate} \cdot \prob_\ingrp^\timeidx \\
-\frac{\win \cdot \npop_\ingrp \cdot \parrate \cdot (\mean_\ingrp^\timeidx)^{\parrate-1}}{\npop_\ingrp \cdot (\mean_\ingrp^\timeidx)^\parrate + \npop_\outgrp \cdot (\mean_\outgrp^\timeidx)^\parrate} \cdot \prob_\outgrp^\timeidx
& 1 - \leak + \frac{\win \cdot \parrate \cdot (\mean_\outgrp^\timeidx)^{\parrate-1}}{\npop_\ingrp \cdot (\mean_\ingrp^\timeidx)^\parrate + \npop_\outgrp \cdot (\mean_\outgrp^\timeidx)^\parrate} \cdot \prob_\ingrp^\timeidx
\end{pmatrix}
\end{align*}

Now, let us consider the undesirable equilibrium case, where $\mean_\ingrp^* = \epsilon$ with $\epsilon \approx 0$%
\footnote{A mean of zero would imply an ill-defined Pareto density. Therefore, we consider here a mean close to zero.},
$\mean_\outgrp^* = \frac{\win}{\leak} \cdot \frac{\naccpt}{\npop_\outgrp}$, $\prob_\ingrp^* = 0$, and
$\prob_\outgrp^* = \frac{\naccpt}{\npop_\outgrp}$. For this equilibrium, we obtain the following Jacobian.

\begin{equation*}
\jac(\epsilon, \frac{\win}{\leak} \cdot \frac{\naccpt}{\npop_\outgrp})
\approx
\begin{pmatrix}
1 - \leak & 0 \\
0 & 1 - \leak
\end{pmatrix}
\end{equation*}
The two eigenvalues of this matrix are both $1 - \leak$. Therefore, for a nonzero leak rate, the
absolute value of these eigenvalues is smaller than one and therefore stability theory implies that the
equilibrium is stable.

Next, let us consider the desirable equilibrium case, where $\mean^* = \mean_\ingrp^* = \mean_\outgrp^* =
\frac{\win}{\leak} \cdot \frac{\naccpt}{\npop}$ and $\prob^* = \prob_\ingrp^* = \prob_\outgrp^* = \frac{\naccpt}{\npop}$.
For this equilibrium, we obtain the following Jacobian.
\begin{align*}
\jac(\frac{\win}{\leak} \cdot \frac{\naccpt}{\npop}, \frac{\win}{\leak} \cdot \frac{\naccpt}{\npop})
&= \begin{pmatrix}
1 - \leak + \frac{\win \cdot \parrate \cdot (\mean^*)^{\parrate-1}}{\npop \cdot (\mean^*)^\parrate} \cdot \prob^*
& -\frac{\win \cdot \npop_\outgrp \cdot \parrate \cdot (\mean^*)^{\parrate-1}}{\npop \cdot (\mean^*)^\parrate} \cdot \prob^* \\
-\frac{\win \cdot \npop_\ingrp \cdot \parrate \cdot (\mean^*)^{\parrate-1}}{\npop \cdot (\mean^*)^\parrate} \cdot \prob^*
& 1 - \leak + \frac{\win \cdot \parrate \cdot (\mean^*)^{\parrate-1}}{\npop \cdot (\mean^*)^\parrate} \cdot \prob^*
\end{pmatrix} \\
&= \begin{pmatrix}
1 - \leak + \frac{\win \cdot \parrate}{\npop} \cdot \frac{\prob^*}{\mean^*}
& -\win \cdot \parrate \cdot \frac{\npop_\outgrp}{\npop} \cdot \frac{\prob^*}{\mean^*} \\
-\win \cdot \parrate \cdot \frac{\npop_\ingrp}{\npop} \cdot \frac{\prob^*}{\mean^*}
& 1 - \leak + \frac{\win \cdot \parrate}{\npop} \cdot \frac{\prob^*}{\mean^*}
\end{pmatrix} \\
&= \begin{pmatrix}
1 - \leak \cdot \big[1 + \frac{\parrate}{\npop}\big]
& -\leak \cdot \parrate \cdot \frac{\npop_\outgrp}{\npop} \\
-\leak \cdot \parrate \cdot \frac{\npop_\ingrp}{\npop}
& 1 - \leak \cdot \big[1 + \frac{\parrate}{\npop}\big]
\end{pmatrix}
\end{align*}
The eigenvalues of this Jacobian are $\eig_1 = 1 - \leak \cdot \big[1 + \frac{\parrate}{\npop} \cdot (1 + \sqrt{\npop_\ingrp \cdot \npop_\outgrp})\big]$
and $\eig_1 = 1 - \leak \cdot \big[1 + \frac{\parrate}{\npop} \cdot (1 + \sqrt{\npop_\ingrp \cdot \npop_\outgrp})\big]$.
The absolute values of these eigenvalues exceed $1$ if the following respective conditions hold.
\begin{align*}
\eig_1 > 1 \quad \iff \quad & \parrate > \frac{\npop}{\sqrt{\npop_\ingrp \cdot \npop_\outgrp} - 1} \\
\eig_2 < -1 \quad \iff \quad & \parrate > \big(\frac{2}{\leak} - 1\big) \cdot \frac{\npop}{\sqrt{\npop_\ingrp \cdot \npop_\outgrp} + 1}
\end{align*}
The former condition can be fulfilled easily if $\parrate > 2$ and $\npop_\ingrp \approx \npop_\outgrp \approx \frac{\npop}{2}$.
As such, this equilibrium is unstable for a wide range of possible settings.

In summary, we have demonstrated that the Pareto distribution yields attractive equilibria in
undesirable positions, whereas the desirable equilibria are unstable for a wide range of conditions.

\subsection{Stability for the Gaussian Distribution}
\label{sec:gaussian}

In the Gaussian distribution model, we assume that the densities
$\dens_\ingrp^\timeidx$ and $\dens_\outgrp^\timeidx$
have the following form:
\begin{align*}
\dens_\ingrp^\timeidx(\quali) &= \nDens(\quali|\mean_\ingrp^\timeidx, \nDev) = \frac{1}{\sqrt{2\pi \nDev^2}} \cdot \exp\Big( - \frac{1}{2} \cdot \frac{(\quali - \mean_\ingrp^\timeidx)^2}{\nDev^2}\Big) \\
\dens_\outgrp^\timeidx(\quali) &= \nDens(\quali|\mean_\outgrp^\timeidx, \nDev) = \frac{1}{\sqrt{2\pi \nDev^2}} \cdot \exp\Big( - \frac{1}{2} \cdot \frac{(\quali - \mean_\outgrp^\timeidx)^2}{\nDev^2}\Big)
\end{align*}
where $\nDens$ denotes the Gaussian density function and where $\sigma$ is the standard deviation of the Gaussian,
which we assume to be fixed and equal across groups.

From these densities we obtain the following success probabilities $\prob_\ingrp^\timeidx$ and
$\prob_\outgrp^\timeidx$.
\begin{align*}
\prob_\ingrp^\timeidx &= \int_{\thresh^\timeidx}^\infty \dens_\ingrp^\timeidx(\quali) d\quali =
1 - \nCum\Big(\frac{\thresh^\timeidx - \mean_\ingrp^\timeidx}{\nDev}\Big)\\
\prob_\outgrp^\timeidx &= \int_{\thresh^\timeidx}^\infty \dens_\outgrp^\timeidx(\quali) d\quali = 1 - \nCum\Big(\frac{\thresh^\timeidx - \mean_\outgrp^\timeidx}{\nDev}\Big)
\end{align*}
where $\nCum$ is the cumulative density function of the standard Gaussian distribution.

Now, let us consider the undesirable equilibrium case, where $\mean_\ingrp^* = 0$ and $\mean_\outgrp^*
= \frac{\win}{\leak} \cdot \frac{\naccpt}{\npop_\outgrp}$. First, note that the threshold $\thresh^\timeidx$
is lower-bounded in this condition due to Equation~\ref{eq:threshold}.
In particular, we obtain the following lower bound.
\begin{align*}
&&1 - \nCum\big(\frac{\thresh^\timeidx - \mean_\outgrp^*}{\nDev}\big)
&\leq \frac{\naccpt}{\npop_\outgrp} \\
\iff && \frac{\thresh^\timeidx - \mean_\outgrp^*}{\nDev} &\geq \nCum^{-1}\big(1 - \frac{\naccpt}{\npop_\outgrp}\big) \\
\iff && \thresh^\timeidx &\geq \nDev \cdot \nCum^{-1}\big(1 - \frac{\naccpt}{\npop_\outgrp}\big) + \mean_\outgrp^*
\end{align*}
Note that this term is strictly larger than zero.
Accordingly, for sufficiently small $\nDev$ and small $\mean_\ingrp^\timeidx<\mean_\outgrp^*$ we obtain:
\begin{equation*}
\prob_\ingrp^\timeidx = 1 - \nCum\Big(\frac{\thresh^\timeidx - \mean_\ingrp^\timeidx}{\nDev}\Big)
\leq 1 - \nCum\Big(\nCum^{-1}\big(1 - \frac{\naccpt}{\npop_\outgrp}\big) + \frac{\mean_\outgrp^* - \mean_\ingrp^\timeidx}{\nDev}\Big)
\leq 1 - \nCum\Big(\frac{\mean_\outgrp^* - \mean_\ingrp^\timeidx}{\nDev}\Big)
\approx 1 - 1 = 0
\end{equation*}
In other words, for sufficiently small $\nDev$ we obtain $\prob_\ingrp^\timeidx \approx \prob_\ingrp^* = 0$,
even if we vary $\mean_\ingrp^\timeidx$ slightly. Accordingly, $\prob_\outgrp^\timeidx =
\frac{\naccpt - \npop_\ingrp \cdot \prob_\ingrp^*}{\npop_\outgrp} \approx \frac{\naccpt}{\npop_\outgrp} = \prob_\outgrp^*$.
This results in the following Jacobian matrix of Equation~\ref{eq:general_model} for small
$\mean_\ingrp^\timeidx$, $\mean_\outgrp^\timeidx \approx \mean_\outgrp^*$ and small $\nDev$.
\begin{equation*}
\jac(\mean_\ingrp^\timeidx, \mean_\outgrp^\timeidx) =
\begin{pmatrix}
\deriv{\mean_\ingrp^\timeidx} \big[(1 - \leak) \cdot \mean_\ingrp^\timeidx + \win \cdot 0\big] &
\deriv{\mean_\outgrp^\timeidx} \big[(1 - \leak) \cdot \mean_\ingrp^\timeidx + \win \cdot 0\big] \\
\deriv{\mean_\ingrp^\timeidx} \big[(1 - \leak) \cdot \mean_\outgrp^\timeidx + \win \cdot \frac{\naccpt}{\npop_\outgrp}\big] &
\deriv{\mean_\outgrp^\timeidx} \big[(1 - \leak) \cdot \mean_\outgrp^\timeidx + \win \cdot \frac{\naccpt}{\npop_\outgrp}\big]
\end{pmatrix}
=
\begin{pmatrix}
1 - \leak & 0 \\
0 & 1 - \leak
\end{pmatrix}
\end{equation*}
The two eigenvalues of this matrix are both $1 - \leak$. Therefore, for a nonzero leak rate, the
absolute value of these eigenvalues is smaller than one and therefore stability theory implies that the
equilibrium is stable.

Next, let us consider the desirable equilibrium case, where $\mean_\ingrp^* = \mean_\outgrp^* =
\frac{\win}{\leak} \cdot \frac{\naccpt}{\npop}$ and $\prob_\ingrp^* = \prob_\outgrp^* = \frac{\naccpt}{\npop}$.
In this equilibrium condition, we can obtain the threshold $\thresh^*$ as follows.
\begin{align*}
&&\frac{\naccpt}{\npop} &= \prob_\ingrp^* = 1 - \nCum\big(\frac{\thresh^* - \mean_\ingrp^*}{\nDev}\big) \\
\iff && \frac{\thresh^* - \mean_\ingrp^*}{\nDev} &= \nCum^{-1}\big(1 - \frac{\naccpt}{\npop}\big) \\
\iff && \thresh^* &= \nCum^{-1}\big(1 - \frac{\naccpt}{\npop}\big) \cdot \nDev + \mean_\ingrp^*
\end{align*}
Now, let us consider small deviations of $\mean_\ingrp^\timeidx$ and $\mean_\outgrp^\timeidx$
which are such that the threshold $\thresh^\timeidx$ stays equal to $\thresh^*$. In that case,
we obtain the following Jacobian of our model in Equation~\ref{eq:general_model}.
\begin{align*}
\jac(\mean_\ingrp^\timeidx, \mean_\outgrp^\timeidx) &=
\begin{pmatrix}
\deriv{\mean_\ingrp^\timeidx} \big[(1 - \leak) \cdot \mean_\ingrp^\timeidx + \win \cdot \big(1 - \nCum(\frac{\thresh^* - \mean_\ingrp^\timeidx}{\nDev})\big)\big] &
\deriv{\mean_\outgrp^\timeidx} \big[(1 - \leak) \cdot \mean_\ingrp^\timeidx + \win \cdot \big(1 - \nCum(\frac{\thresh^* - \mean_\ingrp^\timeidx}{\nDev})\big)\big] \\
\deriv{\mean_\ingrp^\timeidx} \big[(1 - \leak) \cdot \mean_\outgrp^\timeidx + \win \cdot \big(1 - \nCum(\frac{\thresh^* - \mean_\outgrp^\timeidx}{\nDev})\big)\big] &
\deriv{\mean_\outgrp^\timeidx} \big[(1 - \leak) \cdot \mean_\outgrp^\timeidx + \win \cdot \big(1 - \nCum(\frac{\thresh^* - \mean_\outgrp^\timeidx}{\nDev})\big)\big]
\end{pmatrix} \\
&=
\begin{pmatrix}
1 - \leak - \win \cdot \nDens(\thresh^*|\mean_\ingrp^\timeidx, \nDev) & 0 \\
0 & 1 - \leak - \win \cdot \nDens(\thresh^*|\mean_\outgrp^\timeidx, \nDev)
\end{pmatrix}
\end{align*}
Accordingly, the Jacobian at our equilibrium is given as follows.
\begin{align*}
\jac(\mean_\ingrp^*, \mean_\outgrp^*)
&= \begin{pmatrix}
1 - \leak - \win \cdot \nDens(\thresh^*|\mean_\ingrp^*, \nDev) & 0 \\
0 & 1 - \leak - \win \cdot \nDens(\thresh^*|\mean_\outgrp^*, \nDev)
\end{pmatrix} \\
&= \begin{pmatrix}
1 - \leak - \win \cdot \nDens\Big(\nCum^{-1}\big(1 - \frac{\naccpt}{\npop}\big) \cdot \nDev\Big|0, \nDev\Big) & 0 \\
0 & 1 - \leak - \win \cdot \nDens\Big(\nCum^{-1}\big(1 - \frac{\naccpt}{\npop}\big) \cdot \nDev\Big|0, \nDev\Big)
\end{pmatrix} \\
&= \begin{pmatrix}
1 - \leak - \frac{\win}{\nDev} \cdot \nDens\Big(\nCum^{-1}\big(1 - \frac{\naccpt}{\npop}\big)\Big|0, 1\Big) & 0 \\
0 & 1 - \leak - \frac{\win}{\nDev} \cdot \nDens\Big(\nCum^{-1}\big(1 - \frac{\naccpt}{\npop}\big) \Big|0, 1\Big)
\end{pmatrix}
\end{align*}
The two eigenvalues of this Jacobian are both $1 - \leak - \frac{\win}{\nDev} \cdot \nDens\Big(\nCum^{-1}\big(1 - \frac{\naccpt}{\npop}\big)\Big|0, 1\Big)$.
Accordingly, our equilibrium is unstable if $|1 - \leak - \frac{\win}{\nDev} \cdot \nDens\Big(\nCum^{-1}\big(1 - \frac{\naccpt}{\npop}\big)\Big|0, 1\Big)| > 1$.
Given that $\leak$, $\win$, and the Gaussian density function are all non-negative, we can re-write the instability condition as follows.
\begin{align*}
&&-1 &> 1 - \leak - \frac{\win}{\nDev} \cdot \nDens\Big(\nCum^{-1}\big(1 - \frac{\naccpt}{\npop}\big)\Big|0, 1\Big) \\
\iff && \alpha - 2 &> - \frac{\win}{\nDev} \cdot \nDens\Big(\nCum^{-1}\big(1 - \frac{\naccpt}{\npop}\big)\Big|0, 1\Big) \\
\iff && \nDev &<  \frac{\win}{2 - \leak} \cdot \nDens\Big(\nCum^{-1}\big(1 - \frac{\naccpt}{\npop}\big)\Big|0, 1\Big)
\end{align*}
In other words, for sufficiently small $\nDev$, the equilibrium is unstable.

In summary, we have demonstrated that the Gaussian distribution yields attractive equilibria in
undesirable positions, whereas the desirable equilibria are unstable for a wide range of conditions.

\end{appendix}

\end{document}